\documentclass[twoside,11pt]{article}

\usepackage{jmlr2e}
\usepackage{float}
\usepackage{amsmath}

\jmlrheading{}{}{}{}{}{Harsh Patel and Shivam Sahni}

\firstpageno{1}

\begin{document}
\title{Exploring Explainability Methods for Graph Neural Networks}
\author{\name Harsh Patel \email harsh.patel@iitgn.ac.in \\
\addr Computer Science and Engineering, IIT Gandhinagar\\
\name Shivam Sahni 
\email shivam.sahni@iitgn.ac.in\\
\addr Computer Science and Engineering, IIT Gandhinagar
}
\maketitle

\begin{abstract}
With the growing use of deep learning methods, particularly graph neural networks, which encode intricate interconnectedness information, for a variety of real tasks, there is a necessity for explainability in such settings. In this paper, we demonstrate the applicability of popular explainability approaches on Graph Attention Networks (GAT) for a graph-based super-pixel image classification task. We assess the qualitative and quantitative performance of these techniques on three different datasets and describe our findings.  The results shed a fresh light on the notion of explainability in GNNs, particularly GATs.
\end{abstract}


\section{Introduction}
\begin{figure}[!ht]
    \centering
    \includegraphics[width=\linewidth]{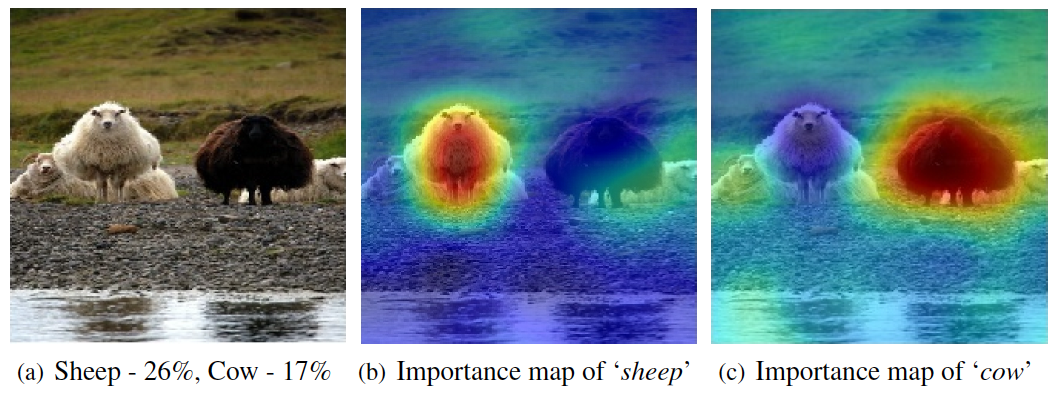}
    \caption{CNN explainability visualisation from \cite{https://doi.org/10.48550/arxiv.1806.07421}}
    \label{cnn_explain}
\end{figure}

Deep Neural Networks have been criticized for being black boxes of information. To explain the results generated by these models, a variety of techniques have been used to enhance their interpretability. For example in Figure \ref{cnn_explain} (b),(c), the heat maps show the explainability results on a Convolutional Neural Network (CNN), ResNet-50, which performs a classification task (\cite{https://doi.org/10.48550/arxiv.1806.07421}). As can be seen from the importance map of cow, the model confuses the black sheep for a cow and thus mispredicts its presence in the image. Such answers, obtained using explainability methods, can be useful in rectifying the mistakes of a model.  
Graph Neural Networks (GNNs) are deep learning models which have proven to be a powerful tool for exploiting graphical information. They do so by combining the node feature information with the graph structure by passing messages along the edges of the graph which makes GNN a complex model. GNN explainability can increase trust in GNNs, improve transparency of model's working and allow users to identify and improve GNN's mistakes.

Traditional convolutional networks use kernels that are limited to domains which use rectangular grids such as 2D images. However, the more complicated sources of visualisations like panoramas capture 360 degree view of a place. Additionally, point cloud classification relies on spatially unstructured data which again cannot be represented through a rectangular grid. We can exploit GNNs to solve the above two problems. 
The use of graphs allows us to model images using pixel level representation or super pixel level representations of images, which are a group of pixels that share common characteristics (like pixel intensity). Superpixels have a perceptual meaning since pixels belonging to a given superpixel share similar visual properties. They provide a convenient and compact representation of images that can be very useful for computationally demanding problems. 

In this paper, we choose to perform superpixel image classification with a broader aim to provide explainability to the GNN predictions. We then compare the explainability results on varying the number of superpixels in the image. Finally, we look at a way to measure these explainability results known as Fidelity.

\section{Related Work}

The difficulty in interpretation and explanation of performance of deep neural networks has been a long standing problem (\cite{Pope_Kolouri_Rostami_Martin_Hoffmann_2019}). Several explainability methods have been developed for deep neural networks, specially CNNs (\cite{Simonyan_Vedaldi_Zisserman_2014}, \cite{Selvaraju_Cogswell_Das_Vedantam_Parikh_Batra_2020}). Methods like Gradient-weighted Class Activation Mapping (Grad-CAM), and Excitation Back-Propagation (EB) have proven to be quite effective on CNNs for image classification tasks. More recently, the work (\cite{Pope_Kolouri_Rostami_Martin_Hoffmann_2019}) demonstrates tools for explaining the performance of Graph Convolutional Neural Networks (GCNNs) on visual scene graphs and molecular graphs.
By considering each pixel as a node, graph-based approaches can be directly applied to images. Lower-level image representation (e.g., Super-pixel segmentation) can, on the other hand, be more effective in classification tasks due to the smaller graphs that are formed. \cite{https://doi.org/10.48550/arxiv.2002.05544} present a methodology for super-pixel image classification using Graph Attention Networks.

\section{Problem Description}
\begin{figure}[!ht]
    \centering
    \includegraphics[width=\linewidth]{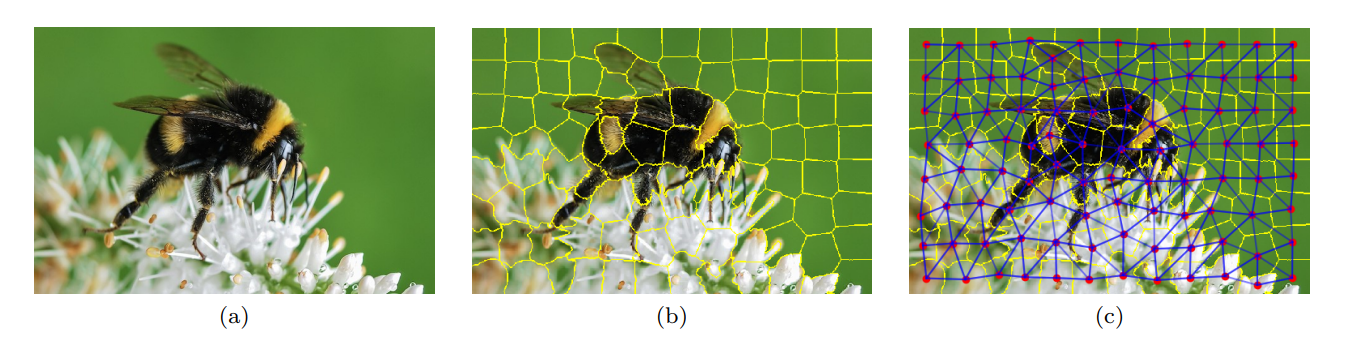}
    \caption{Superpixel graph generation - a) shows the original image, b) shows the superpixels nodes formed using the SLIC algorithm, and c) neighbouring superpixel nodes form an edge which are shown in blue and the nodes are shown in red}
    \label{superpixel}
\end{figure}

A superpixel is a collection of pixels that share similar features like color, pixel intensity, etc. These superpixels give a simple and compact visual representation of an image without losing its perceptual meaning. Many fundamental tasks in computer vision can be applied to these over-segmented and simplified images, including image classification (\cite{Long_yan_chen_2021}). These superpixels can be generated using several approaches like Simple Linear Iterative Clustering (SLIC, \cite{6205760}), Superpixels Extracted via Energy-Driven Sampling (SEEDS, \cite{DBLP:journals/corr/BerghBRG13}). We use the SLIC algorithm for this task which uses an iterative approach to perform the segments for a desired number of equally sized superpixels. Figure \ref{superpixel}(b) shows the superpixeled version of the image \ref{superpixel}(a). This super-pixeled image is further formulated into a graph structure by a simple region adjacency graph generation approach. Each superpixel segment is representative of a node in the graph and all the neighbouring segments form the edges with this node. Figure \ref{superpixel}(c) shows the generated graph for the image. This entire graph acts as an input to our GNN model which performs the task of superpixel image classification.

\section{Dataset}
We use the following three standard classification datasets in our experimentation. Each of these datasets have 50,000 training, 10,000 testing examples and include 10 classes.
\vspace{-\topsep}
\begin{itemize}
    \setlength{\parskip}{0pt}
    \setlength{\itemsep}{0pt plus 1pt}
    \item MNIST (\cite{Lecun_Bottou_Bengio_Haffner_1998})
    \item Fashion-MNIST (\cite{Xiao_Rasul_Vollgraf_2017})
    \item CIFAR-10 (\cite{Krizhevsky_2009})
\end{itemize}

\section{Model: Multi-headed Graph Attention Network (GAT)}
We use a multi-headed graph attention network to perform the task of super pixel image classification. Similar to a GNN, this model generates contextualised node embeddings based on the information that its constitutes (Adjacency matrix, Node Features, Edge Features of the input graph). Graph attention network is a generalised version of a Graph Convolution Network (\cite{Kipf_Welling_2017}), which allows the graphs nodes to generate their embeddings by aggregating information from each of their neighbors.

\begin{figure}[!ht]
    \centering
    \includegraphics[scale=0.7]{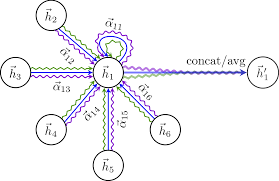}
    \caption{Attention mechanism on node 1 by its neighbourhood (\cite{Velickovic2018GraphAN})}
    \label{gat_example}
\end{figure}

GATs expand this basic aggregation method by using self-attention mechanisms (\cite{Vaswani_Shazeer_Parmar_Uszkoreit_Jones_Gomez_Kaiser_Polosukhin_2017}) that learn the relative importance of each neighbor's contribution, thus improving its learning capacity. An attention coefficient is calculated for each edge in the input graph, so as to improve the aggregation step. Also, similar to the \cite{Vaswani_Shazeer_Parmar_Uszkoreit_Jones_Gomez_Kaiser_Polosukhin_2017} transformer architecture, multi-headed attentions are used in the Graph attention networks for a stabilised learning. Figure \ref{gat_example} illustrates how the attention mechanism aggregates the contribution of each neighbor for the node embedding $\vec{h}_{1}$.

\begin{figure}[!ht]
    \centering
    \includegraphics[scale=0.7]{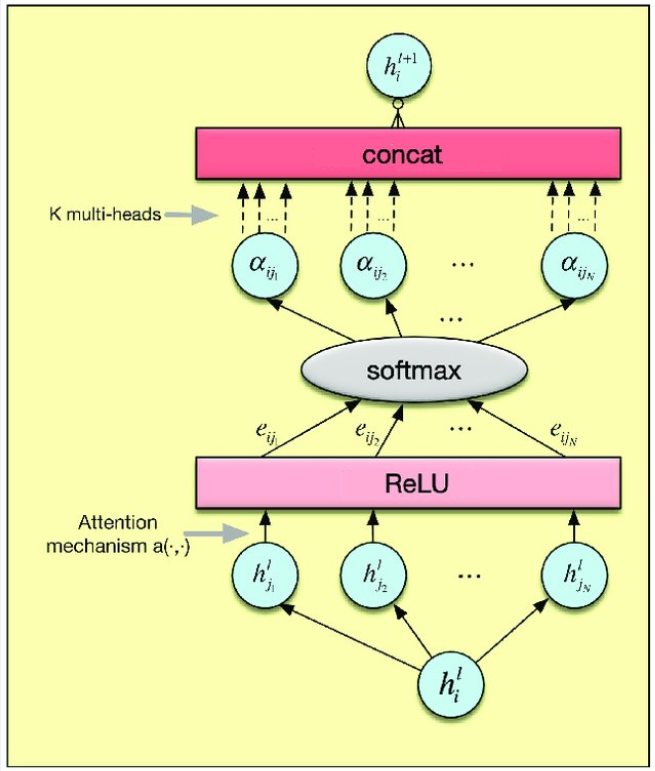}
    \caption{GAT Layer, figure inspired from \cite{article}}
    \label{gat_layer}
\end{figure}

The input to our GAT layer (Figure\ref{gat_layer}) is a set of node features $h$, where N is the
number of nodes, and $F$ is the number of features in each node.
\begin{center}
$
h = \left\{\vec{h}_{1}, \vec{h}_{2}, \ldots, \vec{h}_{N}\right\}, \vec{h}_{i} \in \mathbb{R}^{F}
$
\end{center}
A linear layer is then used to transform the input features into high-level features using a weight matrix $W$. Now, for each edge ($i,j$) in our input graph, we use a linear attention mechanism ($a$) to compute its attention coefficient ($e_{i j}$) that represents its importance in the message passing step of this graph neural network. The input to this attention mechanism is the concatenation of the node embedding of the involved nodes in the edge.
\begin{center}
$e_{i j}^{(l)} = ReLU\left(\vec{a}^{(l)^{T}}(W^{(l)} h_{i}^{(l)} \| W^{(l)} h_{j}^{(l)})\right), W^{(l)} \in \mathbb{R}^{F'\times F}$
\end{center}
These attention coefficients, for each node $i$, are normalised using a softmax operation across all its neighbors ($j$s).
\begin{center}
$\alpha_{i j}^{(l)} = \frac{\exp \left(e_{i j}^{(l)}\right)}{\sum_{k \in \mathcal{N}(i)} \exp \left(e_{i k}^{(l)}\right)}$
\end{center}
The output features for each node are then aggregated using a weighted linear combination of its neighbors' features (the weight being the calculated attention coefficients). In case of multiple heads, we concatenate the scaled attention scores obtained from each independent attention mechanism to generate the final output embeddings. 
\begin{center}
$
h_{i}^{(l+1)} = \vert\vert_{k=1}^{K}\sigma\left(\sum_{j \in \mathcal{N}(i)} \alpha_{i j}^{(l)} z_{j}^{(l)}\right)
$
\end{center}
We thus obtain highly contextualised embeddings for each node in the graph which can be further be passed through feed-forward layers to achieve the task of classification.

\section{Experimental Settings: Training and Evaluation}
\begin{figure}[!ht]
    \centering
    \includegraphics[width=0.7\linewidth]{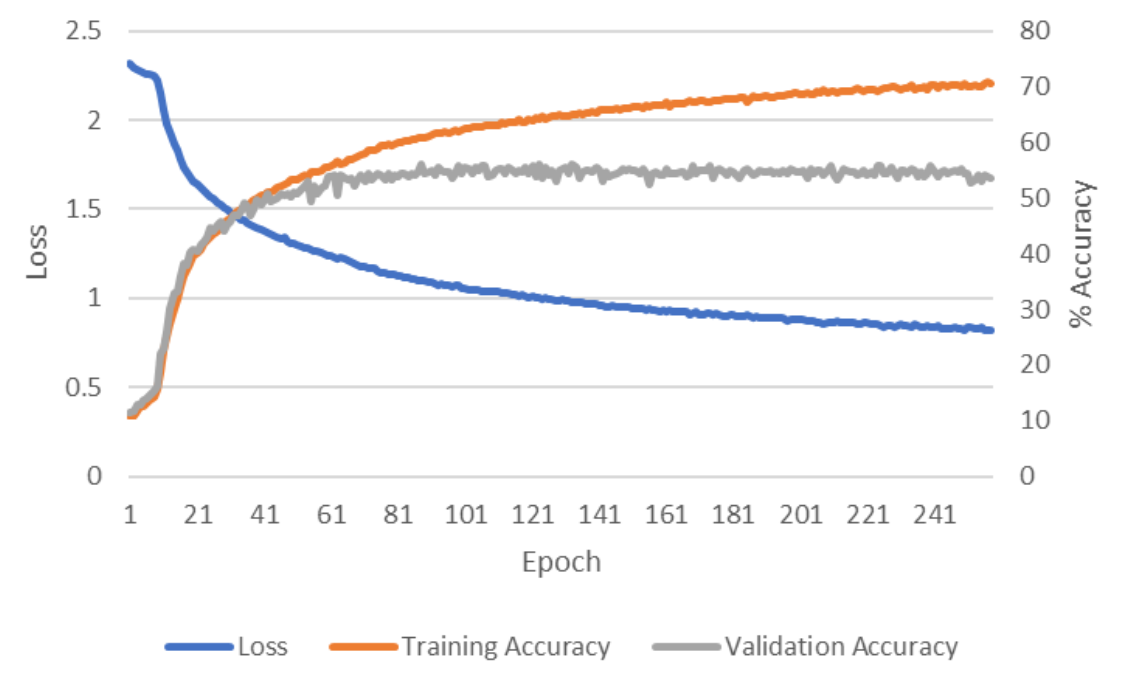}
    \caption{Training curves on CIFAR-10 dataset}
    \label{training}
\end{figure}

For each dataset, we set the number of desired super-pixel segments to be $75$ and train them on a three-headed Graph Attention Network. All experiments are conducted on GPUs. The training on CIFAR-10 dataset for 250 epochs takes around 10 days. Figure \ref{training} shows the variations in classification loss, training and validation accuracy with number of epochs. Table \ref{tab:accuracies} shows the testing accuracies obtained on the three datasets using their corresponding trained models. We use these trained models to perform a variety of explainability methods in an attempt to retrieve useful interpretations and insights on their performance and learning behavior.
\begin{table}[]
    \centering
    \begin{tabular}{c|c}
         \textbf{Dataset} & \textbf{Accuracy (\%)}  \\
         \hline
          CIFAR-10 & 49.73 \\
          MNIST (pretrained) & 97.8 \\
          Fashion MNIST (pretrained) & 89.6\\
    \end{tabular}
    \caption{Model Accuracies}
    \label{tab:accuracies}
\end{table}
\section {Explainability Methods}
We utilise four different explainability methods which were originally intended to work on CNN and can be extended to GCN (\cite{Pope_Kolouri_Rostami_Martin_Hoffmann_2019}). The methods are Contrastive Gradient-based Saliency Maps (CGSM) \cite{Simonyan_Vedaldi_Zisserman_2014}, Class Activation Mapping (CAM) \cite{https://doi.org/10.48550/arxiv.1512.04150}, Grad-Class Activation Mapping (Grad-CAM) \cite{Selvaraju_Cogswell_Das_Vedantam_Parikh_Batra_2020}, Guided-Backpropagation (GBP) \cite{https://doi.org/10.48550/arxiv.1412.6806}.
\subsection{Contrastive Gradient-based Saliency Maps (CGSM)}
CGSM is the most straightforward way to implement explainability. This method relies on the derivative of the output of the model with respect to the input image. We apply ReLU to the gradients to discard negative gradients.  

$$L_{\text {Gradient }}^{c}=\left\|\operatorname{ReLU}\left(\frac{\partial y^{c}}{\partial x}\right)\right\|$$
where $y^{c}$ is the score of class $c$ and $x$ is the input. It is argued that CGSM represents noise more than signals.

\subsection{Class Activation Mapping (CAM)}
CAM is based on the motivation that the class features at the last convolutional layer are more meaningful. CAM has an architectural constraint in which it requires the layer immediately before the final softmax classifier to be a convolutional layer followed by a global average pooling layer. Assume, $F_{k}$ to be the $k^{th}$ feature map of the output from last convolutional layer. The global average pooling is defined as:
\[e_{k}=\frac{1}{Z} \sum_{i} \sum_{j} F_{k, i, j}\]
The global average pooled output is then passed to a softmax classifier to get the output class scores.
\[y^{c}=\sum_{k} w_{k}^{c} e_{k},\]
The importance of a pixel in CAM is defined as:
\[
L_{C A M}^{c}[i, j]=\operatorname{ReLU}\left(\sum_{k} w_{k}^{c} F_{k, i, j}\right)
\]
\subsection{Grad-Class Activation Mapping (Grad-CAM)}
Grad-CAM relaxes the assumption the architectural restriction put forward by CAM by replacing the feature map weights with the back propagated gradients. 
\begin{center}
$\alpha_{k}^{c}=\frac{1}{Z} \sum_{i} \sum_{j} \frac{\partial y^{c}}{\partial F_{k, i, j}}$
\end{center}
\begin{center}
$L_{G r a d-C A M}^{c}[i, j]=\operatorname{ReLU}\left(\sum_{k} \alpha_{k}^{c} F_{k, i, j}\right)$
\end{center}
\subsection{Guided-Backpropagation (GBP)}
\begin{figure}[!ht]
    \centering
    \includegraphics[width=0.8\linewidth]{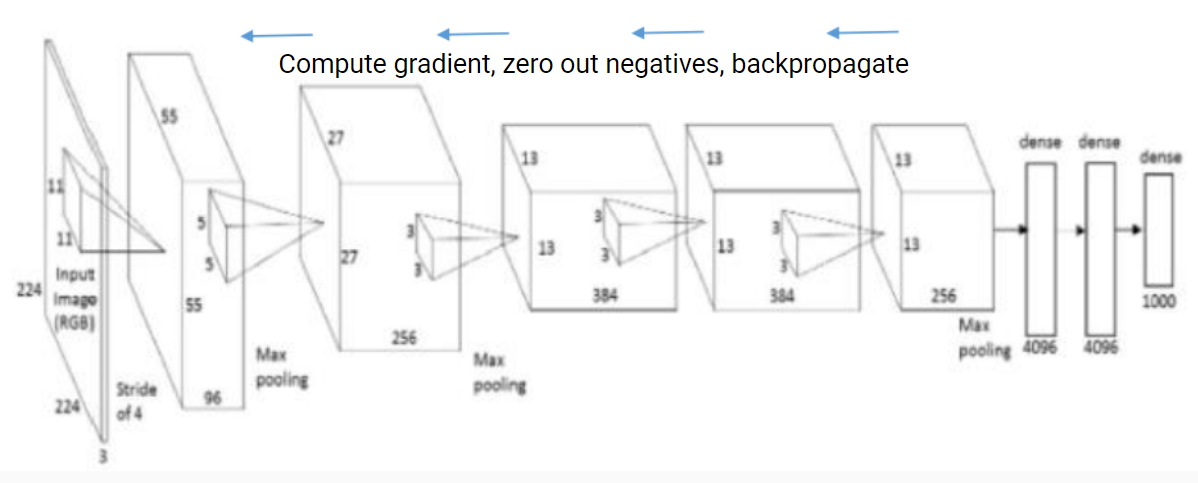}
    \caption{Intuitive explanation of guided backpropagation (GBP)}
    \label{gbp}
\end{figure}

The motivation behind Guided-Backpropagation is to discard neagtive gradients during backpropagation as negative gradients play the role of suppressing the output. As shown in Fig \ref{gbp} we compute gradients, then zero out the negative ones and continue to backpropagate.
\label{guidedbp}

\section {Explainability Quantification: Fidelity}
We use the Fidelity score as the quantification metric for all the explanaiblity methods. The key idea behind Fedality is that the occlusion of important features, discovered through the obtained explanations, on the original image would lead to a reduction in classification accuracy. It is defined as the difference in accuracy obtained by excluding all nodes with a saliency value greater than 0.01 (on a scale 0 to 1).
\begin{figure}[!ht]
    \centering
    \includegraphics[width=0.7\linewidth]{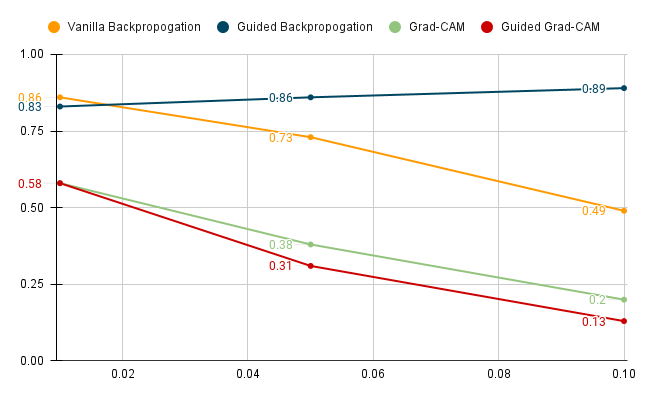}
    \caption{Fidelity metric results on the four explainability methods by varying occlusion threshold}
    \label{fidelity}
\end{figure}
Figure \ref{fidelity} shows the fidelity scores of each of the above discussed explainability methods with varying threshold on the MNIST dataset. We observe that out of all the 4 methods, Guided-Backpropagation (Section \ref{guidedbp}) shows the best fidelity, i.e. occluding its strongly identified regions from the image, leads to higher number of misclassifications. We also observe that on increasing the threshold (i.e., occluding lesser number of nodes), the fidelity score decreases because occluding less number of nodes would generally lead to a very less change in the classification accuracy. However, in Guided-Backpropagation, the fidelity scores do not decrease with increasing threshold. The gradients in this method are backpropogated only through the relevant neurons, which means that the contributing pixels have a direct and significant correlation with the classification output. Thus, even on occluding lesser number of nodes, the misclassifications are still observed because of the high saliency values of the occluded nodes.

\section {Results}
In this section, we show explainability results on the three datasets and provide a qualitative analysis of the results obtained.
\begin{figure}[!ht]
    \centering
    \includegraphics[width=\linewidth]{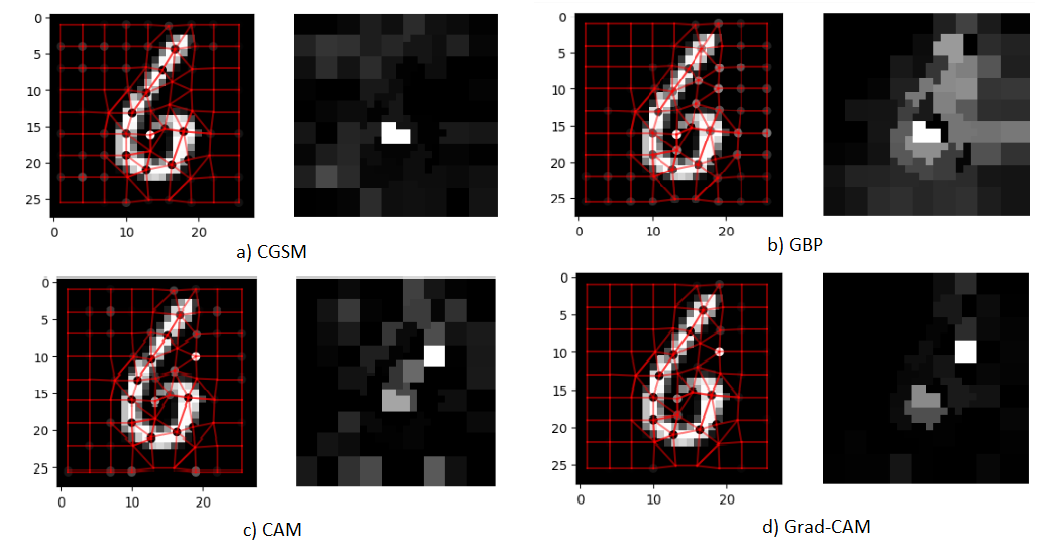}
    \caption{Visualizations of the four explainability methods on `6' from MNIST dataset}
    \label{viz-6}
\end{figure}
The figure \ref{viz-6} shows the visualisations from all the four methods on an example from MNIST dataset representing the number `6'. We observe that Guided Backpropagation seems to capture all the nodes involved in making the number look like the number `6', and hence performs the best. We also observe that the Vanilla Backpropagation method has the maximum noise and the Guided Grad-CAM has the minimum noise, however they both fail to fully capture all the important nodes.

\begin{figure}[!ht]
    \centering
    \includegraphics[width=\linewidth]{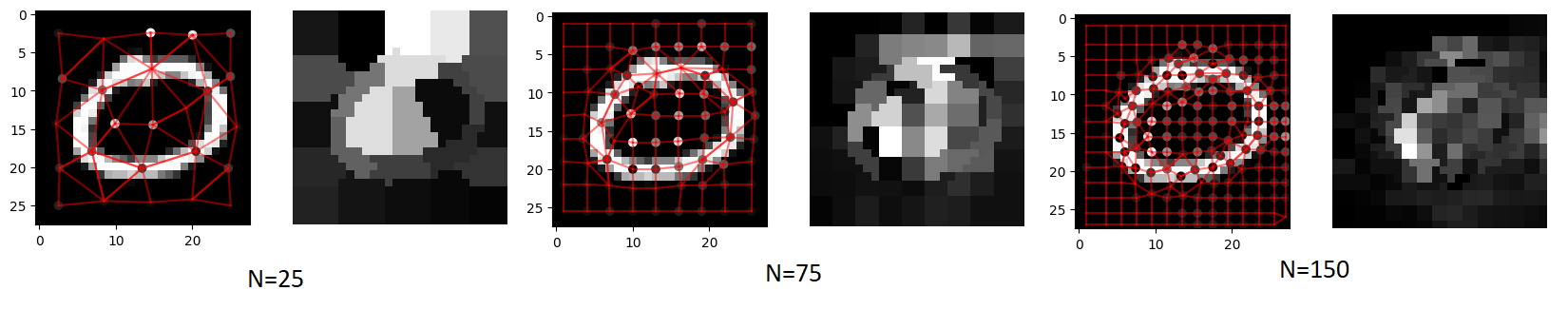}
    \caption{Visualizations}
    \caption{Visualization of GBP by varying the number of superpixels on `0' from MNIST dataset}
    \label{viz-0}
\end{figure}
The figure \ref{viz-0} shows the visualisations by varying the number of superpixels in the image. From the figure we observe that as the total number of superpixel increases from 25-150, the total amount of region enclosed outside `0' decreases, thus we can say that the noise in prediction decreases. 

\begin{figure}[!ht]
    \centering
    \includegraphics[]{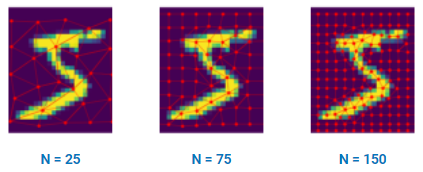}
    \caption{Variation in the superpixel graph on varying the number of desired superpixels}
    \label{nodesV_example}
\end{figure}
\begin{figure}[!ht]
    \centering
    \includegraphics[]{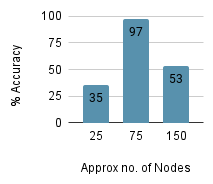}
    \caption{Accuracy vs Varying no. of superpixels}
    \label{nodesV}
\end{figure}

Figure \ref{nodesV_example}
shows an exmaple image from a MNIST dataset super-pixeled with varying number of segments. We also compare the \% classification accuracy of the GAT model with varying number of nodes in the input graph. We observe that using a very few number of nodes leads to loss of information which leads to the classification accuracy of 35\% on the MNIST dataset, whereas increasing the number of nodes to 150, leads to increased complexity in message passing and also leads to the formation of super pixels that do not carry forward a perceptual meaning. This again leads to a decrease in the overall accuracy.

\begin{figure}[!ht]
    \centering
    \includegraphics[width=\linewidth]{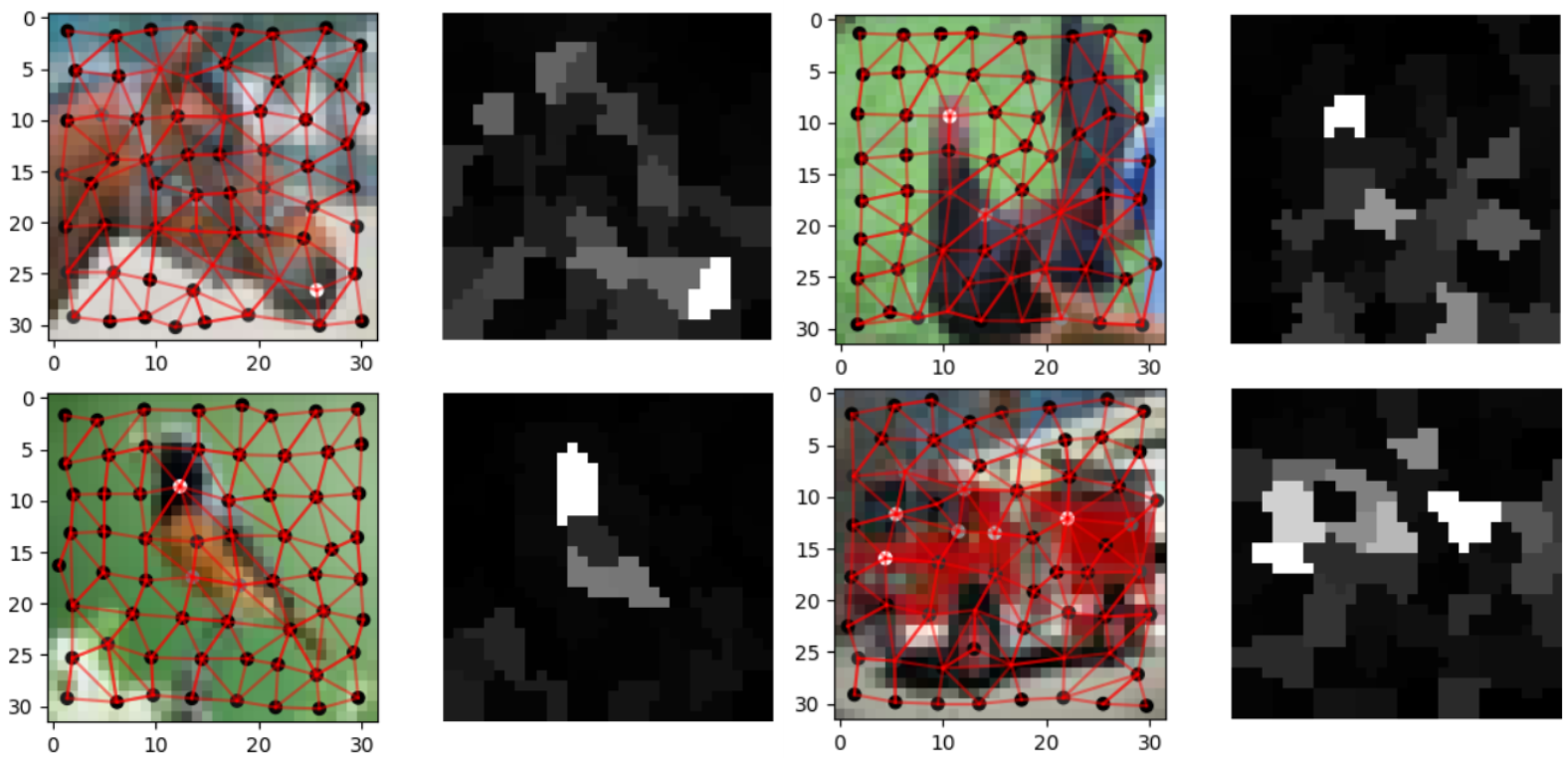}
    \caption{Visualizations Results on CIFAR 10 dataset}
    \label{viz-good}
\end{figure}
The figure \ref{viz-good} shows some explainability results on the CIFAR-10 dataset. We observe that the silhouette of the explainability visualizations match the objects depicted in the image. This provides assurance on the quality of the explainability visualizations we get.

\section{Conclusion}
In this paper, we performed superpixel image classification task on three standard classification datasets and visualized various explainability methods in Graph Neural Network settings, particularly Graph Attention Networks (GATs). We characterized the explanations with the fidelity (quantification metric) and found that Guided-Backpropagation method shows the best performance amongst all the methods. We also provide performance analysis of the explainability methods with varying number of superpixels in the images during the classification task.


\newpage








\vskip 0.2in
\bibliography{sample}

\end{document}